\documentclass[runningheads]{llncs}
\usepackage[T1]{fontenc}
\usepackage{multirow}
\usepackage{graphicx}
\usepackage{siunitx}
\usepackage{graphicx,amsmath,hyperref,color,tikz,pgfplots,subfig}
\pgfplotsset{compat=1.18}

\begin{document}
\title{Safe Road-Crossing by Autonomous Wheelchairs: a Novel Dataset and its Experimental Evaluation}
\titlerunning{Safe Road-Crossing by Autonomous Wheelchairs}
\author{Carlo Grigioni \orcidID{0009-0002-0794-0612} \and
Franca Corradini\orcidID{0009-0009-5867-9478} \and
Alessandro Antonucci\orcidID{0000-0001-7915-2768}
\and J\'er\^ome Guzzi\orcidID{0000-0002-1263-4110}
\and Francesco Flammini\orcidID{0000-0002-2833-7196}
\institute{IDSIA USI-SUPSI, University of Applied Sciences and Arts of Southern Switzerland, Lugano, Switzerland}\\
\email{name.surname@supsi.ch}}
\authorrunning{C. Grigioni et al.}
\maketitle
\begin{abstract}
Safe road-crossing by self-driving vehicles is a crucial problem to address in smart-cities. In this paper, we introduce a multi-sensor fusion approach to support road-crossing decisions in a system composed by an autonomous wheelchair and a flying drone featuring a robust sensory system made of diverse and redundant components. 
To that aim, we designed an analytical danger function based on explainable physical conditions evaluated by single sensors, including those using machine learning and artificial vision. As a proof-of-concept, we provide an experimental evaluation in a laboratory environment, showing the advantages of using multiple sensors, which can improve decision accuracy and effectively support safety assessment. We made the dataset available to the scientific community for further experimentation. The work has been developed in the context of an European project named REXASI-PRO, which aims to develop trustworthy artificial intelligence for social navigation of people with reduced mobility.
\end{abstract}

\section{Introduction}\label{sec:intro}
Self-driving vehicles have attracted a huge interest in the intelligent transportation systems industry. Those vehicles are equipped with advanced sensors providing data to computing systems running \emph{Artificial Intelligence} (AI) algorithms that enable them to operate and navigate without human intervention. They offer the promise of increased safety and performance. These vehicles rely on several technologies, including LiDARs, radars, cameras, and sophisticated software to perceive their surroundings, make decisions, and control their movements \cite{autonomous_vehicles}. However, all those technological innovations based on AI and \emph{Machine Learning} (ML) come with big challenges in terms of trustworthiness, due to limited transparency and failures possibly having fatal consequences in safety-critical systems \cite{9979717}. This is also the case of \emph{Autonomous Wheelchairs} (AWs) that are meant to support motion-impaired persons in safe door-to-door navigation.

There have been several efforts to improve the technology of powered wheelchairs since their first appearance during the 19th century~\cite{yukselir2012brains}. Modern AWs can be very complex including sophisticated components, such as autonomous navigation system, and aim to accommodate several disabilities by means of multi-modal interfaces. Safety assessment and certifications of AWs are paramount aspects to address \cite{tas23}. In fact, in presence of components based on AI, together with specific regulations such as ``ISO 7176-14:2022 Wheelchairs'',\footnote{\url{https://www.iso.org/standard/72408.html}} other requirements and guidelines must be considered such as the ones included in the EU Artificial Intelligence Act.\footnote{\url{https://artificialintelligenceact.eu}}

The work described in this paper has been developed within the European project REXASI-PRO (REliable \& eXplAinable Swarm Intelligence for People with Reduced mObility),\footnote{\href{https://rexasi-pro.spindoxlabs.com/}{https://rexasi-pro.spindoxlabs.com/}} which addresses indoor and outdoor use cases to demonstrate trustworthy social navigation of autonomous wheelchairs together with flying drones in real-world environments. Drones are used to provide a view of the environment from different perspectives and thus collect additional data 
to scan and map the surrounding environment. 

In this paper, we deal with a road-crossing scenario using multiple sensors and with no support of traffic lights. We adopt multi-sensor fusion for run-time risk-based decision support. To this aim, we use laboratory wheeled robots, equipped with artificial vision and distance sensors, as mock-up models of actual AWs and drones, for which we perform run-time safety evaluation using an analytical danger function based on physical considerations.

The main original contributions of this paper can be summarized as follows:
\begin{itemize}
    \item We provide a novel reference scenario for road-crossing by AWs as well as hints and details about laboratory experimentation and performance evaluation.
    \item We share with the scientific community a specific dataset 
    with data recorded from multiple sensors that can be used for further experimentation and performance evaluation.
    \item We introduce an approach for road-crossing decision making based on an analytical and hence explainable danger function highlighting the importance of multi-sensor obstacle detection. 
    \item As a proof-of-concept, we report the results of a preliminary experimental evaluation demonstrating the advantages of using diverse redundant sensors, which can support safety certification when used in combination with probabilistic safety models.
\end{itemize}

The remaining of this paper is structured as follows. Sec.~\ref{sec:lit} provides an overview of the related literature. Sec.~\ref{sec: road-crossing} introduces the reference road-crossing scenario. Sec.~\ref{sec:df} describes the design of the danger function. Sec.~\ref{sec:data_gen } addresses dataset generation in the laboratory environment. Sec.~\ref{sec:exp} reports the results of the experimental evaluation. Finally, Sec.~\ref{sec:conclusion} provides conclusions and hints about future developments.

\section{Related works}\label{sec:lit}

Pedestrians and wheelchair users are among the most vulnerable group type in transport systems and traffic environments \cite{Pedestrian,Obstacles_wheel,statistics2019analysis}. Being road-crossing especially dangerous, it is one of the most investigated situation in terms of pedestrian–vehicle conflict to prevent accidents \cite{ELHAMDANI2020102856}.
%
In \cite{Crosswalk}, authors propose a crosswalk safety warning system where both pedestrian and vehicle behaviour is detected and evaluated by multiple sensors; warning alerts are communicated to both parts through lights integrated in the cross path infrastructure. In \cite{10010764}, a smart traffic light control system equipped with a camera is used to manage pedestrian crossing considering the traffic condition and type of pedestrian; crossing is improved for the elderly or people with disabilities without worsening traffic.
In \cite{9043590}, authors propose a vision-based approach to generate safety messages in real-time using video streams from roadside traffic cameras; the messages can be used by connected autonomous vehicles and smart phones equipped with pedestrian safety applications.
Given the costs of such approaches, virtual reality is a useful tool to simulate smart pedestrian crossing \cite{guo2023smart}, particularly if designed for wheelchair users \cite{Co-Designing-VR}.


Although those solutions represent an aid for wheelchair users, they need a specific and possibly expensive infrastructure.
In \cite{Wheel_video}, ML is used to elaborate the videos of on-board cameras and to support AW decisions. A similar approach is applied in \cite{INDRA}, where a convolutional neural network is suitably designed to evaluate the risk of road-crossing on Indian roads. Compared to our approach, those works are based on single ML technologies lacking sufficient redundancy, diversity and transparency, and as such less suitable for safety assessment.



\begin{figure}
  \centering
  \begin{minipage}{1\linewidth}
    \centering
    \subfloat{\includegraphics[width=.32\linewidth]{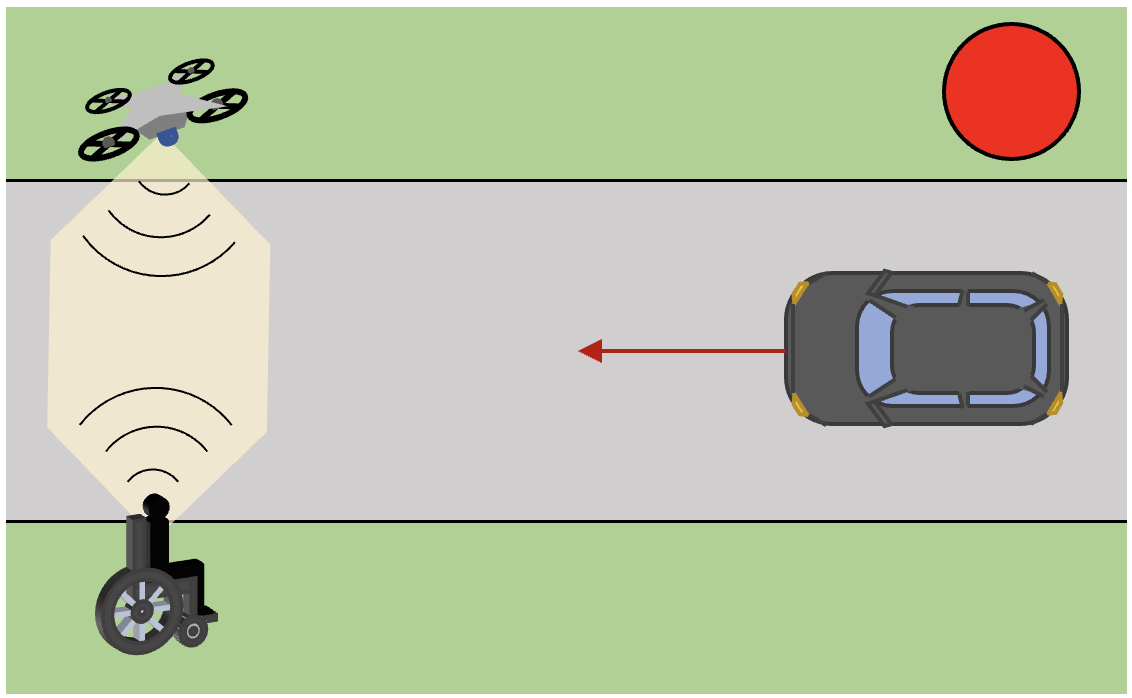}}
    \subfloat{\includegraphics[width=.32\linewidth]{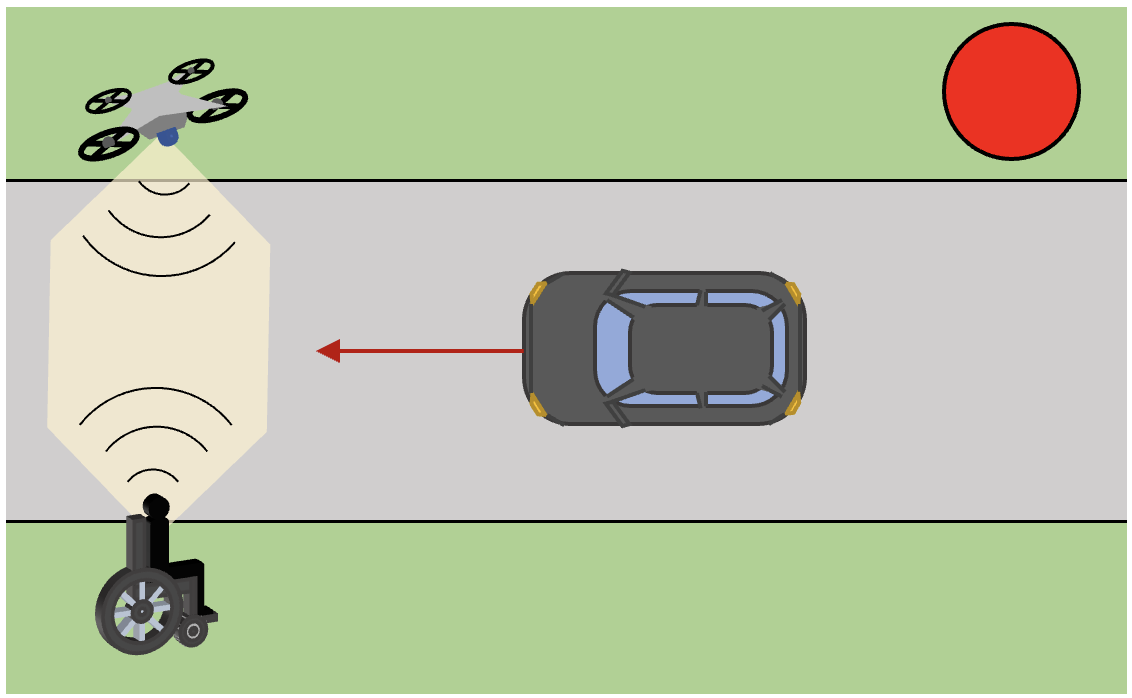}}
    \subfloat{\includegraphics[width=.32\linewidth]{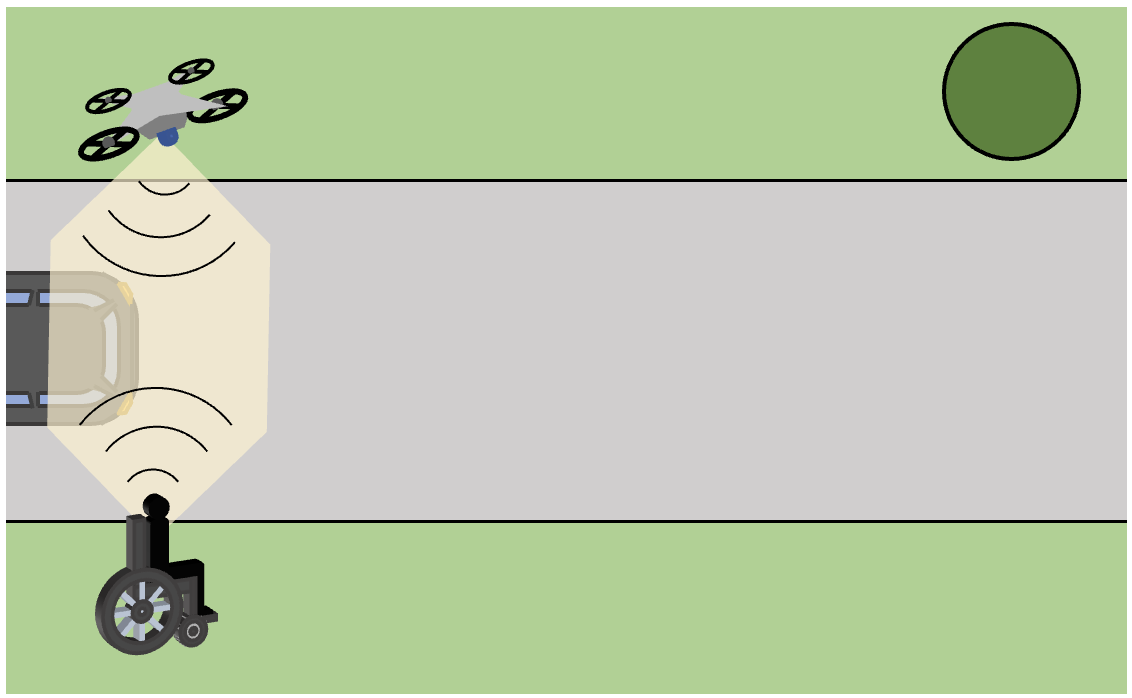}}

    \caption*{(a)}
  \end{minipage}\quad
  \begin{minipage}{1\linewidth}
    \centering
    \subfloat{\includegraphics[width=.32\linewidth]{images/scene1.png}}
    \subfloat{\includegraphics[width=.32\linewidth]{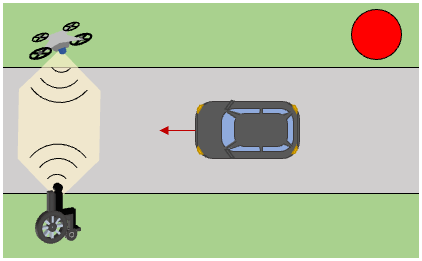}}
    \subfloat{\includegraphics[width=.32\linewidth]{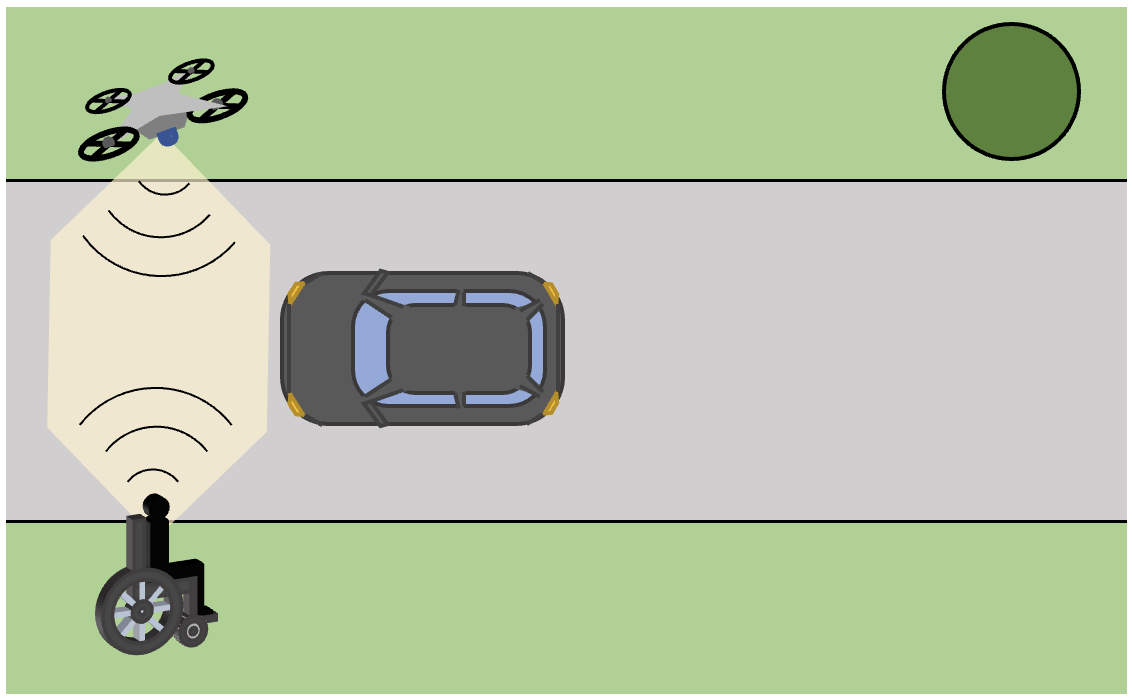}}
    \caption*{(b)}
  \end{minipage}

\caption{Road-crossing scenario: drone and AW cooperate to support safe decisions, e.g., when a car surpasses the crossing location (a), or when a car stops before the crossing location (b).}
\label{fig:syntet_scen}
\end{figure}

\section{Reference scenario for safe road-crossing}\label{sec: road-crossing}

When crossing a road, especially where there are no traffic lights, humans consider a variety of factors to ensure both safety and convenience. Primarily, they assess the flow of traffic, considering variables like the speed and quantity of vehicles. They also account for the presence of road-crossings, traffic signals, and zebra crossings, as these demarcate safe areas for traversing. The distance to the nearest crossing point and the estimated time required to reach it are taken into consideration, as are potential traffic delays that might offer a safer crossing opportunity. Additionally, individuals evaluate their own physical capabilities for quick crossing and gauge the visibility of oncoming vehicles. Factors such as weather conditions, the presence of children or elderly, and the accessibility of nearby sidewalks also affect the decision. Ultimately, the act of crossing a road involves a multifaceted interplay of these considerations, striking a balance between personal safety and the need to reach the destination efficiently.

In our analysis, we assume that pedestrians account for distance, speed and acceleration (\cite{PAPIC2020105586} and \cite{acceleration}) as pivotal cues in their assessment of approaching vehicles before determining whether it is safe to cross the road. These factors provide vital insights into how rapidly a vehicle is closing in and whether adequate time exists for a safe crossing.

\paragraph*{Distance.} The distance of an approaching vehicle is a central indicator on which pedestrians heavily rely. Distance perception is related to the personal experience of each single pedestrian. For example it might be affected by the physical condition of the pedestrian, the weather condition or other elements that can affect the time required to perform crossing. 

\paragraph*{Speed.} People gauge speed through visual cues, such as changes in the vehicle's apparent size and the interval it takes for the vehicle to traverse a specific distance between reference points on the road. A vehicle that is swiftly shortening the gap is generally perceived as moving at a higher speed, while one with a more gradual progression may be perceived as moving slower. 

\paragraph*{Acceleration.} Pedestrians also consider acceleration, denoting how swiftly an object alters its speed over time. Acceleration contributes to the decision-making process by signaling how quickly a vehicle may reach the pedestrian's location. Consequently, this can influence the choice to wait for the vehicle to pass before attempting to cross. 



\hfill \break
We apply a similar reasoning to the road-crossing scenario for AWs, where the system must take a safe road-crossing decision. Following the principles of redundancy and diversity, a safe decision must be based on multiple sensors using different technologies whose accuracy is affected by mostly independent factors \cite{Flammini2020}. This can be achieved, for instance, by combining cameras featuring artificial vision with LiDARs or other distance sensors. Additional information can be obtained by connecting the AW with drones or specific infrastructures \cite{9043590}, providing information from diverse sources or different perspectives. 

The road-crossing scenario we consider consists of an AW on the side of a one-way road aiming to reach the other side, a drone connected to the AW that provides further information and a vehicle approaching the specific section of the road. We assume that there are no traffic lights and the driver’s behaviour is not predictable; he/she might slow down to allow the crossing or move ahead, maintaining its motion. Hence, the AW must use its sensing capabilities to estimate the potential danger: the crossing is safe if the danger is estimated under a certain threshold. If the danger is too high, crossing is not recommended until the vehicle slows down or surpasses the crossing location. This is why it is important to design a danger function, as described in the next section. Fig.~\ref{fig:syntet_scen} provides a schematic representation of the scenario.

\section{Design of the danger function}\label{sec:df}
\begin{figure}[htp!]
    \centering
    \includegraphics[width=9cm]{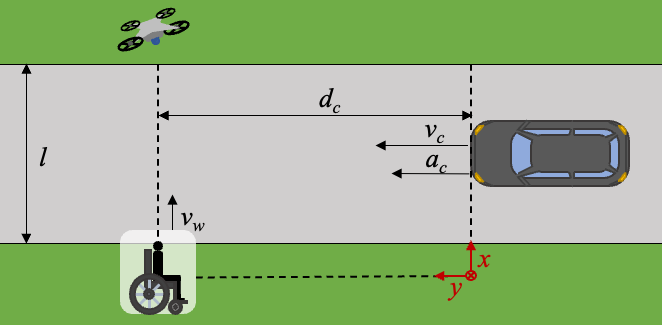}
    \caption{Kinematics involved in the design of the danger function.}
    \label{fig:raod_crossing}
\end{figure}

In this section, we address the problem of assessing the risk associated with road-crossing based on what discussed in the previous section. The road-crossing scenario includes an AW attempting to cross the road and an oncoming obstacle (e.g., a car). As shown in Fig.~\ref{fig:raod_crossing}, the two bodies are assumed to have orthogonal motions, with the pedestrian crossing the road from one side to the other while the vehicle proceeds along the driveway.

In our approach, the risk evaluation is based on a continuous function $g$, called \emph{Danger Function} (DF). The function is intended to reflect a kinematic analysis of the road-crossing scenario in a particular moment $t$. Accordingly, the value of the DF reflects the danger level associated with the kinematic configuration of the system at a specific moment. For decision-making, we summarise the assessments provided by the DF $g$ by tagging them as potentially dangerous if the corresponding DF values exceed a given threshold $g^*$.

To derive the DF, we assume that the AW can proceed with uniform linear motion and the obstacle approaches it with a uniformly accelerated motion. Both assumptions are often unrealistic, but might reflect an over-cautious modelling providing a safer evaluation: in real scenarios, in case of danger, the vehicle would likely slow down and the AW would likely accelerate.

We consequently model the road-crossing scenario as follows. Assuming the AW crossing the road to move on the x-axis and starting the crossing from position $x=0$ with constant speed $v_{\mathit{w}}$, we have:
\begin{equation}\label{eq:wheelchair}
x_{\mathit{w}}(t) = v_{\mathit{w}} \cdot t\,.
\end{equation}
Similarly, assuming the obstacle, i.e., the car, moving on the y-axis with constant acceleration $a_c$ and initial speed $v_c$:
\begin{equation}\label{eq:obstacle}
y_c(t) = v_c \cdot t + \frac{a_c}{2} \cdot t^2\,.
\end{equation}
From Eq.~\eqref{eq:wheelchair}, we can compute the time $t_{\mathrm{cross}}$ required by the AW to cross a road of width $l$, i.e., $t_{\mathrm{cross}}:=l/v_{\mathit{w}}$. During the interval of time $[0,t_{\mathrm{cross}}]$, the obstacle should be far enough to not impact with the AW, i.e., if $d_c$ is the distance of the obstacle: 
\begin{equation}\label{eq:s_min}
d_c > v_c \cdot t_{\mathrm{cross}} + \frac{a_c}{2} \cdot t_{\mathrm{cross}}^2\,,
\end{equation}
and hence:
\begin{equation}\label{eq:constraint}
\frac{\frac{l}{v_{\mathit{w}}}\cdot v_c + \frac{l^2}{2 \cdot v_{\mathit{w}}^2} \cdot a_c}{d_c}<1\,.
\end{equation}

We can regard the constraint in Eq.~\eqref{eq:constraint} as a safety condition for a kinematics-based DF with threshold one. As expected, the danger increases for higher values of speed and acceleration of the car, while decreasing for increasing distances. Assessing all the parameters of such a DF might be critical when coping with different road-crossing scenarios. For this reason, we heuristically define a DF $g$, with similar relations with $d_c$, $v_c$ and $a_c$ as in Eq.~\eqref{eq:constraint}, whose parameters might be reasonably assessed for a generic scenario.

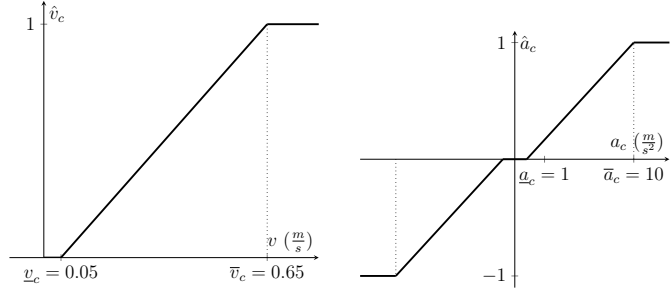
\begin{figure}[htp!]
\centering
\begin{tikzpicture}[scale=0.6]
\begin{axis}[
axis lines=middle,
xmax=0.8,
xmin=-0.1,
ymin=0,
ymax=1.1,
xtick={0.05,0.65}, 
label style = {font=\large},
ticklabel style = {font=\large},
xticklabels={$\underline{v}_c=0.05$,$\overline{v}_c=0.65$},
ytick={0.0,1.0},    
yticklabels={$0$,$1$},
xlabel={$v$  ($\frac{m}{s}$)},
ylabel={$\hat{v}_c$}]
\draw[black,very thick] (0,0) -- (0.05,0);
\draw[black,very thick] (0.05,0) -- (0.65,1);
\draw[black,very thick] (0.65,1) -- (0.8,1);
\draw[dotted] (0.65,0) -- (0.65,1);
\end{axis}
\end{tikzpicture}
\phantom{aa}
\begin{tikzpicture}[scale=0.6]
\begin{axis}[
axis lines=middle,
xmax=13,
xmin=-13,
ymin=-1.1,
ymax=1.1,
xlabel={$a_c$ ($\frac{m}{s^2}$)},
ylabel={$\hat{a}_c$},
ticklabel style = {font=\large},
label style = {font=\large},
xticklabels={$\underline{a}_c=1$, $\overline{a}_c=10$},
xtick={2.5,10},
ytick={-1,0,1}]
\draw[black,very thick] (-13,-1) -- (-10,-1);
\draw[black,very thick] (10,1) -- (13,1);
\draw[black,very thick] (-1,0) -- (1,0);
\draw[black,very thick] (-10,-1) -- (-1,0);
\draw[black,very thick] (10,1) -- (1,0);
\draw[dotted] (10,0) -- (10,1);
\draw[dotted] (-10,0) -- (-10,-1);
\end{axis}
\end{tikzpicture}
\caption{Linear transformations with threshold used for $v_c$ (left) and $a_c$ (right).}
\label{fig:va}
\end{figure}

Regarding the inverse dependence with respect to $d_c$, we assume that close objects are significantly more dangerous than distant objects, even if they are moving at lower speed. Consequently, to enhance the impact of the distance on the function in specific situations where the two bodies are close, a logarithm smoothing is applied to  $d_c$. Moreover, a small threshold $\epsilon$ is also added to always obtain finite values.
Concerning the linear combination of speed and acceleration in the numerator of the left-hand side of Eq.~\eqref{eq:constraint}, we perform linear transformations with threshold on both $v_c$ and $a_c$, and denote the output of these transformations as $\hat{v}_c$ and $\hat{a}_c$ and use a coefficient $k$ to evaluate the relative contribution of the speed with respect to the acceleration.
Therefore the DF is described as:
\begin{equation}\label{eq:complete}
g(d_c, v_c, a_c) := \frac{\hat{v}_c + k \cdot \hat{a}_c}{\log (d_c + \epsilon)}\,,
\end{equation}
where $k=0.1$ and $\epsilon=0.6$ are set following heuristics. The linear transformation of the speed is:
\begin{equation}\label{eq:speed_component}
\hat{v}_c :=
\begin{cases} 
0&\text{if }v_c \leq \underline{v}_c\,,\\
\frac{v_c-\underline{v}_c}{\overline{v}_c-\underline{v}_c}&\text{if }\underline{v}_c< v_c\leq \overline{v}_c\,,\\
1&\text{if }v_c > \overline{v}_c\,,\\
\end{cases}
\end{equation}
and, for the acceleration:
\begin{equation}\label{eq:acceleration_component}
\hat{a}_c:=
\begin{cases} 
-1&\text{if }a_c\leq -\overline{a}_c\\
\frac{a_c+\underline{a}_c}{\overline{a}_c-\underline{a}_c}&\text{if }-\overline{a}_c < a_c \leq -\underline{a}_c\\
0&\text{if }-\underline{a}_c<a_c\leq \underline{a}_c\\
\frac{a_c-\underline{a}}{\overline{a}_c-\underline{a}_c}&\text{if }\underline{a}_c < a_c  \leq \overline{a}_c\\
1 & \text{if }a_c > \overline{a}_c\,.\\
\end{cases}
\end{equation}
The above transformations are intended to prevent contributions to the DF by low speeds (i.e., $v_c\leq \underline{v}_c$) and low accelerations (i.e., $|a_c|\leq \underline{a}_c$), while also putting a normalised upper bound to the contributions of high speeds ($v_c \geq \overline{v}_c$) and accelerations ($|a_c| \geq \overline{a}_c$).
Fig.~\ref{fig:va} shows the choice of the threshold parameters $\underline{v}_c$, $\overline{v}_c$, $\underline{a}_c$, and $\overline{a}_c$, we considered for the experimental setup (Sec.~\ref{sec:exp}).

\begin{figure}[htp!]  
\centering 
\begin{tikzpicture}[scale=0.7,baseline]
\pgfplotsset{xmin=0,xmax=5}
\begin{axis}[axis y line*=left,ymin=0, ymax=3, xlabel=$t$ (s),ylabel=distance $d$ (m)]
\addplot[thick,color=black,mark=none] table [smooth,x=timestamp,y=distance_tracker,col sep=comma]{all.csv};
\label{plot:distance}
\end{axis}
\begin{axis}[axis y line*=right,axis x line=none,ymin=0, ymax=0.6,ylabel=speed $v$ (m/s)]
\addlegendimage{/pgfplots/refstyle=plot:distance}\addlegendentry{distance}
\addplot[thick,dotted,color=black,mark=none] table[smooth,x=timestamp, y=speed_tracker, col sep=comma]{all.csv};
\addlegendentry{speed}
\end{axis}
\end{tikzpicture}
\begin{tikzpicture}[scale=0.7,baseline]
\pgfplotsset{xmin=0,xmax=5}
\begin{axis}[ytick={0.0,1.0,1.5},yticklabels={$0.0$,$g^*$,$1.5$},ymin=0, ymax=1.5,xlabel=$t$ (s),ylabel=danger level $g$]
\addplot[color=black,thick] table[mark=none,smooth,x=timestamp,y=danger_tracker,col sep=comma]{all.csv};
\draw[dotted,line width=0.25mm] (0,1) -- (10,1);
\end{axis}
\end{tikzpicture}
\caption{Distance, speed (left) and DF values (right) on the experiment in Sec.~\ref{sec:exp}.}\label{fig:df_tracker}
\end{figure}
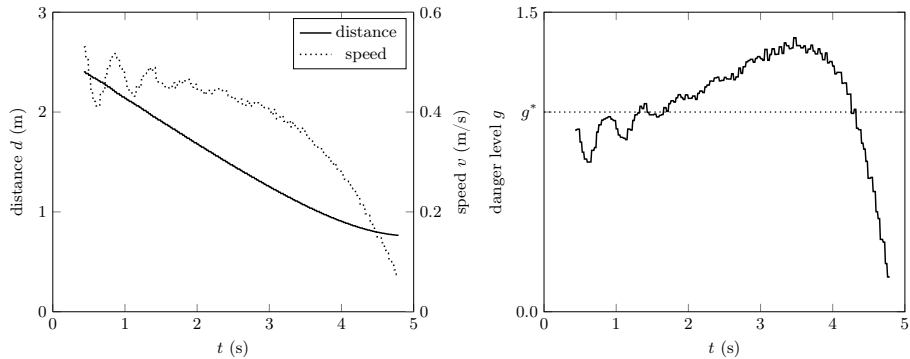

Fig.~\ref{fig:df_tracker} shows an example of the kinematics (for clarity, only distance and speed on the left) and the corresponding values of the DF (right). 

\section{Dataset generation} \label{sec:data_gen }
In this section, we provide the specification of the laboratory environment --- including all the equipment and hardware components --- and describe the procedure of data collection, data pre-processing and data elaboration. In the experiments, we target a similar scenario as described in Sec.~\ref{sec: road-crossing}. 
We adopt a simplified experimental setup, where three ground robots equipped with vision and distance sensors represent the AW, drone and vehicle, as depicted in Fig.~\ref{fig:Scenes}. In our lab, we can access ground truth poses from a very accurate motion tracking system.

\subsection{Laboratory environment}
The experiments have been performed in 
the \emph{IDSIA Autonomous Robotics Laboratory}.\footnote{\href{https://idsia-robotics.github.io}{https://idsia-robotics.github.io}}

\paragraph{Robots.}
For our experiments we use three wheeled omni-directional robots: each one is a RoboMaster EP (RM), a commercial education platform from DJI,\footnote{\href{https://www.dji.com/ch/robomaster-ep}{https://www.dji.com/ch/robomaster-ep}} whose specifications are summarized in Tab.~\ref{tab:Robomaster_spec}.
%
%
%
Each RM is customised for its role as ``car'', ``AW'', or ``drone''.
\begin{description}
\item[RM$_\mathrm{c}$] simulates the vehicle and does not require additional sensors, as we are interested just on its movement.
\item[RM$_{\mathrm{w}}$] represents the AW and is equipped with a camera and four infrared range sensors. Given their narrow \emph{Field Of View} (FOV) (see Tab.~\ref{tab:Robomaster_spec}), range sensors are located on the RM so that their outputs can be combined to obtain information for a larger FOV. The RM$_{\mathrm{w}}$ and its sensors are shown in Fig.~\ref{fig:Scenes}b. Conservatively, we consider the minimum distance reading returned by the four range sensors, which from now on we refer to as the single sensor \emph{Range Sensors Unit} (RSU).
\item[RM$_{\mathrm{d}}$] acts as the drone and has a camera positioned on top of the robotic arm.
\end{description}

\begin{table}[htp]
\centering
\begin{tabular}{lll}
\hline
\multirow{3}{*}{Chassis} & size & \qtyproduct{32 x 24 x 27}{cm}\\
& maximal speed & \SI{3.5}{m/s}\\  
& maximal angular speed& \SI[per-mode=symbol]{600}{\degree\per\second}\\
\hline
\multirow{3}{*}{Camera}  & FOV & \ang{120} \\
& video resolution & 1280 x 720 \\
& video fps & \SI{30}{\Hz} \\
\hline
\multirow{3}{*}{Range sensors} & maximal range & \SI{10}{m} \\
& FOV & \ang{20} \\
& accuracy & \SI{5}{\percent}\\
\hline
\end{tabular}
\caption{Technical specifications of RoboMaster EP.}
\label{tab:Robomaster_spec}
\end{table}

\paragraph{Motion Tracker.}
The lab used for the experiments is equipped with an OptiTrack motion tracker\footnote{\href{https://docs.optitrack.com/v/v2.3}{https://docs.optitrack.com/v/v2.3}} composed of eighteen infrared cameras covering an area of \qtyproduct{6x6x2}{\meter} to track the pose of the three robots at \SI{30}{Hz} with sub-millimetre accuracy. 

\subsection{Data collection and preprocessing} \label{subsec:Data_prepr}
As in the scenario described in Sec.~\ref{sec: road-crossing}, RM$_{\mathrm{w}}$ and RM$_{\mathrm{d}}$ are motionless and oriented towards the approaching RM$_{\mathrm{c}}$, which is remotely controlled. 

\paragraph{Data collection.}
To control the robots and collect data from the motion tracker, we used the \emph{Robot Operating System} (ROS2~\cite{ROS}), executing a ROS2 driver for each robot.\footnote{\href{https://github.com/jeguzzi/robomaster\_ros}{https://github.com/jeguzzi/robomaster\_ros}}
For each experimental run, we recorded data from the two cameras, the RSU and the motion tracker in bag files as detailed in Tab.~\ref{tab:Dataset_spec}.

\begin{figure}[htp!]%
\centering
\subfloat[]{\includegraphics[height=4.5cm]{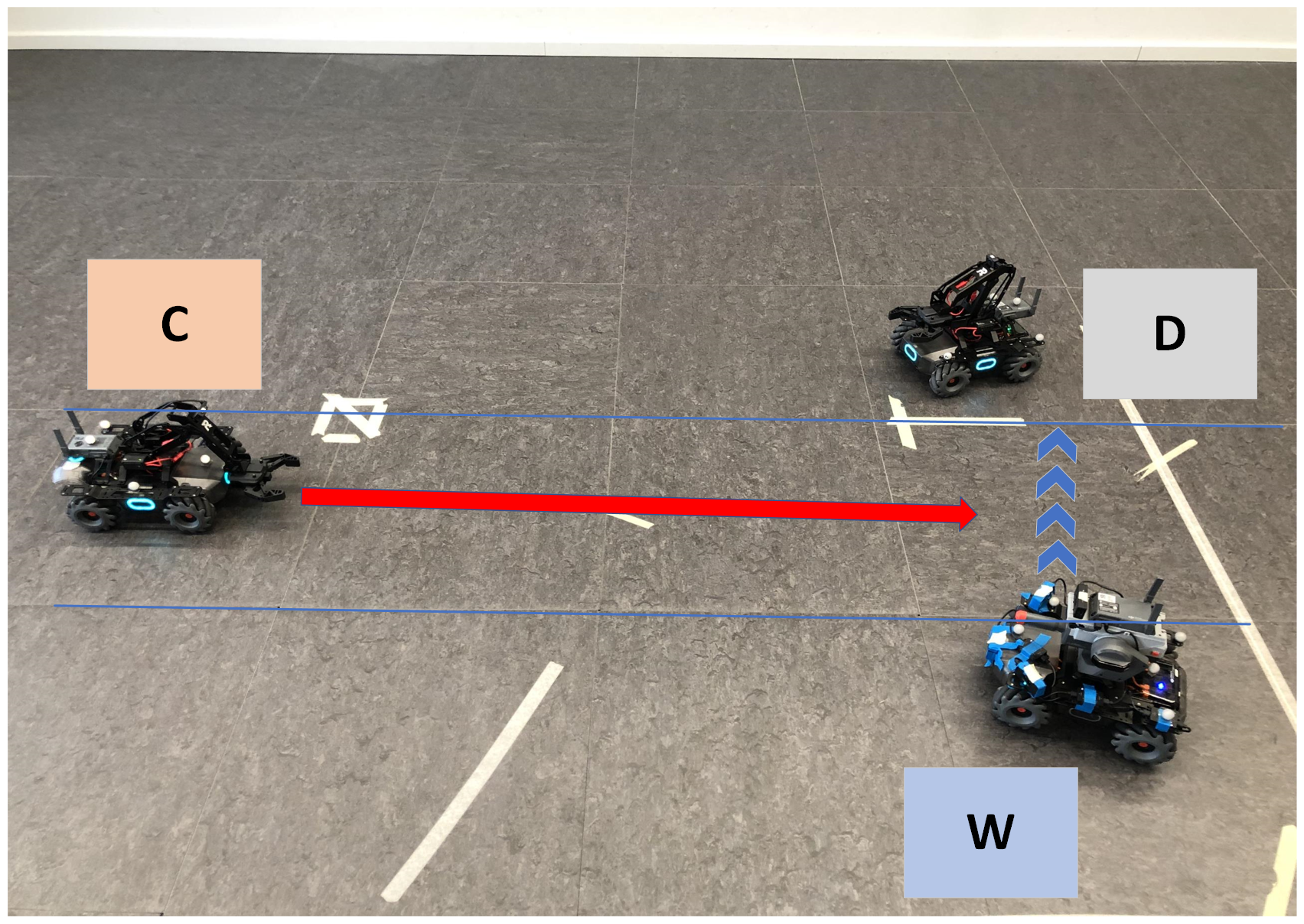}}
\hfill
\subfloat[]{\includegraphics[height=4.5cm]{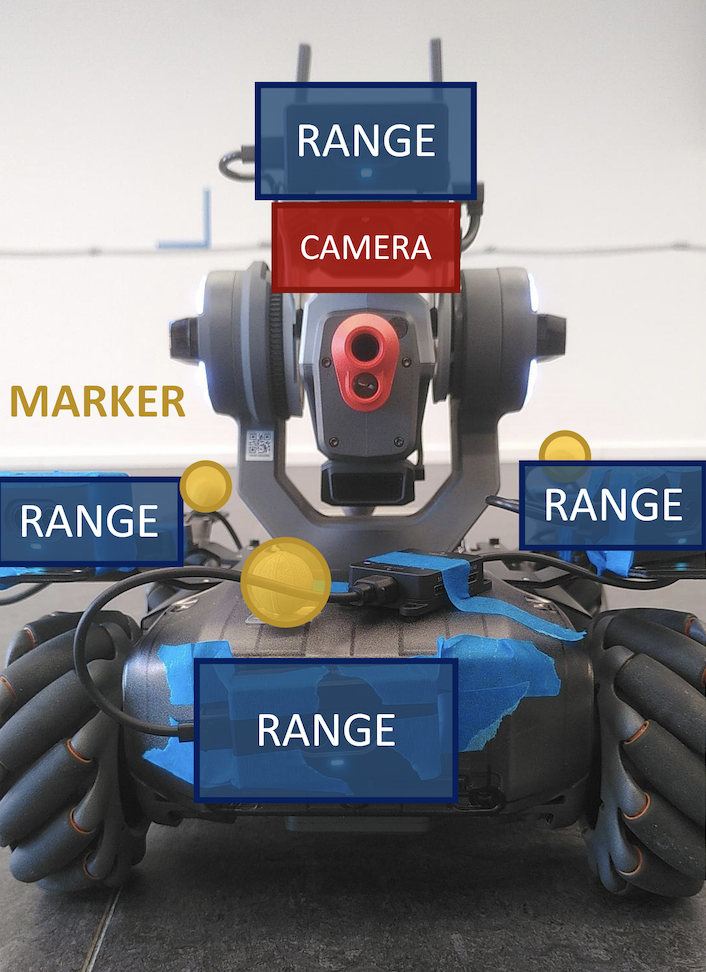}}

\caption{(a) Example of setup for data collection: in red is the trajectory of RM$_{\mathrm{c}}$ and in blue the crossing path of RM$_{\mathrm{w}}$. (b) RM$_{\mathrm{w}}$ and its components.}
\label{fig:Scenes}%
\end{figure}

Overall, we recorded 15 runs for two different setups (the first one is depicted in Fig.~\ref{fig:Scenes}a), for which runs last approximately \SI{6}{s} and resp. \SI{9}{s}. In the second setup, the initial distance of the RM$_{\mathrm{c}}$ with respect to the RM$_{\mathrm{w}}$ is larger, allowing for more complex patterns in movement of the obstacle during experiments.


\begin{table}[htp]
\centering
\begin{tabular}{llll}
\hline
Name & Source & Type & Frequency \\
\hline
Poses of RM$_{\mathrm{w}}$, RM$_{\mathrm{d}}$ and RM$_{\mathrm{c}}$  & Motion Tracker& 2D poses & \SI{30}{Hz}\\
Video & RM$_\mathrm{w}$ & $1080\times720$ RGB images & \SI{30}{Hz} \\
Video & RM$_\mathrm{d}$  & $1080\times720$ RGB images & \SI{30}{Hz} \\
Range sensors & RM$_\mathrm{w}$ & 4 distances to nearest object   & \SI{10}{Hz} \\
\hline
\end{tabular}
\caption{Raw data recorded during the experiments.}
\label{tab:Dataset_spec}
\end{table}

\paragraph{Data pre-processing.}
Raw data from the bag files were processed to synchronise all streams. 
Relative distances between RM$_{\mathrm{w}}$ and RM$_{\mathrm{c}}$ were computed for ground truth poses and from RSU.
%
%
Raw data from cameras were split using OpenCV\footnote{\href{https://docs.opencv.org/4.x/index.html}{https://docs.opencv.org/4.x/index.html}} into frames. These were processed with \emph{You Only Look Once} (YOLO)\footnote{\href{https://github.com/ultralytics/yolov5}{https://github.com/ultralytics/yolov5}} \cite{redmon2016look}, a commonly used Machine Learning model, to detect RM$_{\mathrm{c}}$. RMs do not have a specific class in YOLO, but are consistently recognised as \emph{motorcycles}. YOLO was used to identify their bounding boxes in image space, from which we computed the distance using 
triangle similarity.\footnote{\href{https://pyimagesearch.com/2015/01/19/find-distance-camera-objectmarker-using-python-opencv/}{https://pyimagesearch.com/2015/01/19/find-distance-camera-objectmarker-using-python-opencv/}}
%
%
Finally, the three distance measures from the sensors onboard, i.e. closed RSU and the two cameras, were smoothed, taking an average of the previous 2 (RSU) or 5 (cameras) samples .

\subsection{Data elaboration and sensor fusion} 
\label{subsec:DATA_elab}
\begin{figure}[h]
\centering
\subfloat{\includegraphics[width=.34\linewidth]{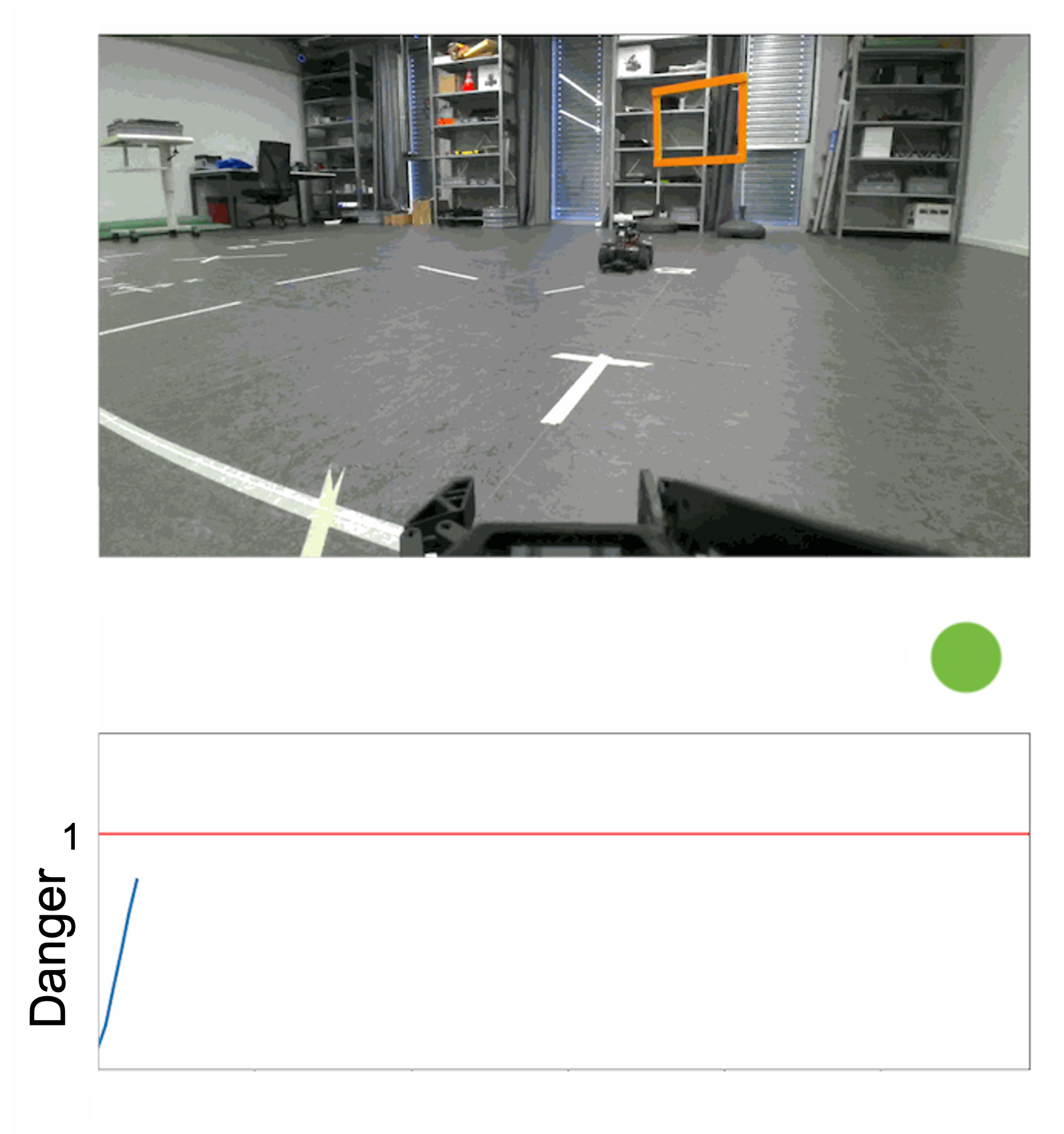}}
\subfloat{\includegraphics[width=.34\linewidth]{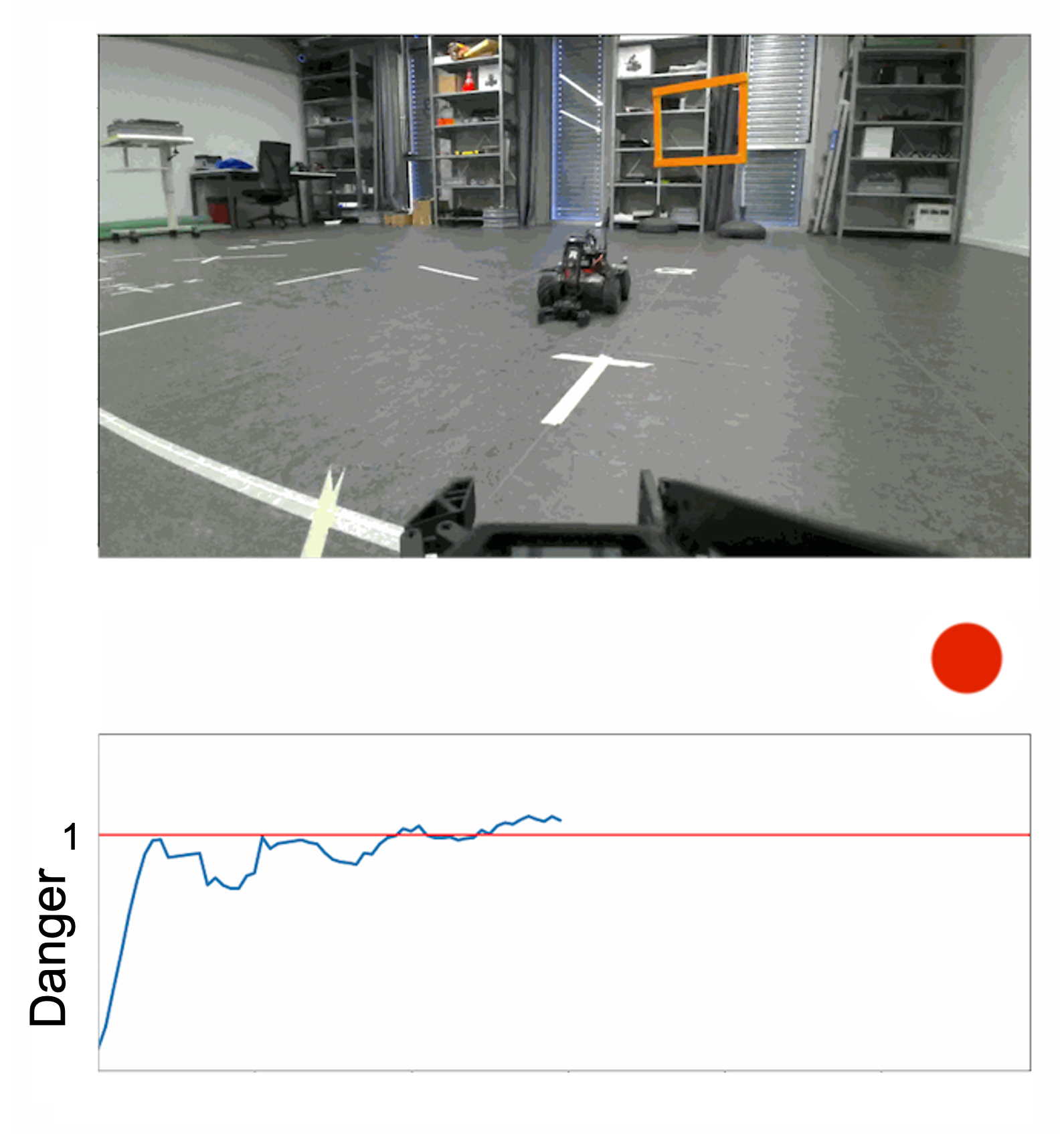}}
\subfloat{\includegraphics[width=.34\linewidth]{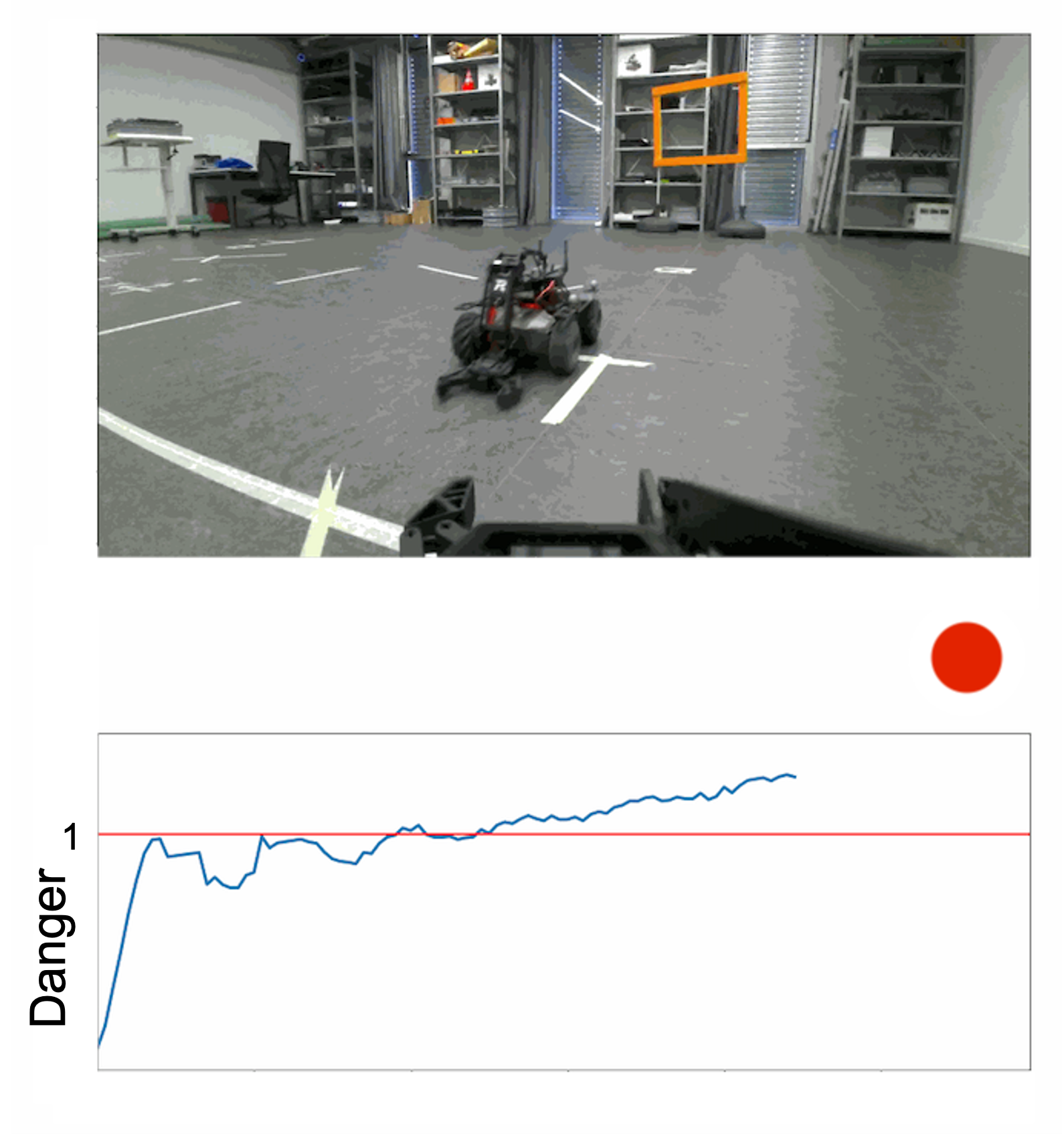}}
\caption{Example of the results: RM$_{\mathrm{w}}$ camera frames number 6, 60 and 90 with corresponding DF calculated by ground truth (tracker) measures.}
\label{fig:video}
\end{figure}

Speed and acceleration measures were computed from the four distance measures obtained from the RSU, the two cameras and the tracker.
Then, danger was estimated for each sensor using the function described in Sec.~\ref{sec:df}. When the danger value exceeded the given threshold, a dangerous situation was detected and crossing was not recommended. An example of the results is shown in Fig.~\ref{fig:video}: three frames from the RM$_{\mathrm{w}}$ camera output are displayed with the corresponding DF value calculated by the tracker. A dot in green (resp. red) defines a safe (resp. dangerous) situation as a result of the threshold criterion.

Since three values can be obtained from the RSU and the two cameras, a fusion procedure is required. 
Raw data from sensors are already aligned, but have different sampling times. 
All measures were therefore resampled at 100 Hz before merging.
We considered three different fusion architectures. A straightforward fusion procedure consists of taking the average of the distances measured by the three sensors. We call this procedure \emph{distance fusion}. We call instead \emph{danger fusion} the fusion procedure where the average of the DFs computed by the distances measured by the three sensors is considered. Finally, \emph{voting fusion} was applied to the values obtained after the threshold operation on the DFs through a majority vote. All the data collected are available, together with the code used for processing and analysis, in a freely available repository.\footnote{\href{https://github.com/CarloGrigioni/safe_roadcrossing_aw}{https://github.com/CarloGrigioni/safe\textunderscore roadcrossing\textunderscore  aw}}.

\begin{figure}[htp!]  
\centering 
\begin{tikzpicture}[baseline]
\pgfplotsset{xmin=0.5,xmax=4.2}
\begin{axis}[
width=.5\textwidth,
height=.22\textheight,
legend style={at={(0.54,0.53)}},
axis y line*=left,ymin=0, ymax=2.5, xlabel=$t$ (s),
xtick={0.0,1.0,2.0,3.0,4.0},ylabel=distance $d$ (m)]
\addplot[thick,color=orange,mark=none] table [smooth,x=timestamp,y=distance_range,col sep=comma]{sensors.csv};
\addlegendentry{\tiny{range sensors}}
\addplot[thick,color=red,mark=none] table [smooth,x=timestamp,y=distance_wheelchair,col sep=comma]{sensors.csv};
\addlegendentry{\tiny{AW camera}}
\addplot[thick,color=blue,mark=none] table [smooth,x=timestamp,y=distance_drone,col sep=comma]{sensors.csv};
\addlegendentry{\tiny{drone camera}}
\addplot[thick,dotted,color=black,mark=none] table [smooth,x=timestamp,y=distance_tracker,col sep=comma]{sensors.csv};
\addlegendentry{\tiny{tracker}}
\end{axis}
\end{tikzpicture}
\begin{tikzpicture}[baseline]
\pgfplotsset{xmin=0.5,xmax=4.2}
\begin{axis}[
width=.5\textwidth,
height=.22\textheight,
legend style={at={(0.65,0.5)}},
axis y line*=left,ymin=0, ymax=2.5, xlabel=$t$ (s),
xtick={0.0,1.0,2.0,3.0,4.0}]
\addplot[very thick,color=black,mark=none] table [smooth,x=timestamp,y=distance_fusion,col sep=comma]{all.csv};
\addlegendentry{\scriptsize{distance fusion}}
\addplot[thick,dotted,color=black,mark=none] table [smooth,x=timestamp,y=distance_tracker,col sep=comma]{all.csv};
\addlegendentry{\scriptsize{tracker}}
\end{axis}
\end{tikzpicture}
\caption{Distance as recorded by the different sensors (left), by the ground truth (tracker)  and after the \emph{distance fusion} (right).}
\label{fig:distance_sensors}
\end{figure}
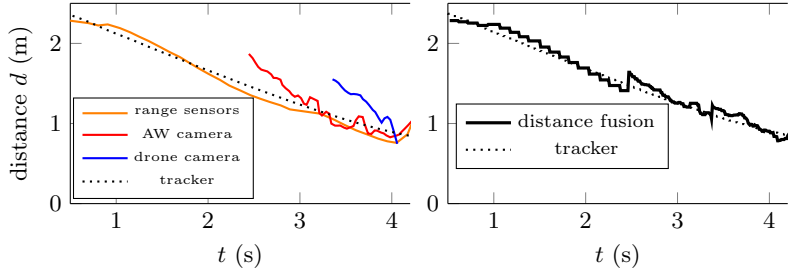

\section{Experimental evaluation}\label{sec:exp}
In this section, we present the results of the three fusion approaches described in Sec.~\ref{subsec:DATA_elab} for one of the experiments. Fig.~\ref{fig:distance_sensors} depicts the comparison between the three pre-processed distance measures with the ``ground truth'' from the motion tracker (left), and the DF based on the \emph{distance fusion} (right).

From Fig.~\ref{fig:distance_sensors} (left), it can be observed that RSU output is always available, while cameras generate signals only when RM$_\mathrm{c}$ is already near to the other RMs. When the output of the cameras is not available,  the \emph{distance fusion} is only computed based on the RSU. This is due to the computer vision system, which is unable to adequately detect the RM$_\mathrm{c}$ since it is confused with the laboratory background when it is far from cameras.

\begin{figure}[htp!]  
\centering 
\begin{tikzpicture}[baseline,scale=0.7]
\pgfplotsset{xmin=0.5,xmax=4.2}
\begin{axis}[
legend style={at={(0.5,0.9)}},
axis y line*=left,ymin=0, ymax=3.5, xlabel=$t$ (s),
xtick={0.0,1.0,2.0,3.0,4.0},ylabel=danger $g$,ytick={1.0},yticklabels={$g^*$}]
\addplot[thick,color=orange,mark=none] table [smooth,x=timestamp,y=danger_range,col sep=comma]{all.csv};
\addlegendentry{\scriptsize{range sensors}}
\addplot[thick,color=red,mark=none] table [smooth,x=timestamp,y=danger_wheelchair,col sep=comma]{all.csv};
\addlegendentry{\scriptsize{AW camera}}
\addplot[thick,color=blue,mark=none] table [smooth,x=timestamp,y=danger_drone,col sep=comma]{all.csv};
\addlegendentry{\scriptsize{drone camera}}
\addplot[very thick,dotted,color=black,mark=none] table [smooth,x=timestamp,y=danger_tracker,col sep=comma]{all.csv};
\addlegendentry{\scriptsize{tracker}}
\draw[color=black!40,line width=0.25mm] (0,1) -- (5,1);
\end{axis}
\end{tikzpicture}
\begin{tikzpicture}[baseline,scale=0.7]
\pgfplotsset{xmin=0.5,xmax=4.2}
\begin{axis}[legend style={at={(0.6,0.9)}},
axis y line*=left,ymin=0, ymax=3.5, xlabel=$t$ (s),
xtick={0.0,1.0,2.0,3.0,4.0},ytick={1.0},yticklabels={$g^*$}]
\addplot[very thick,color=black,mark=none] table [smooth,x=timestamp,y=danger_fusion,col sep=comma]{all.csv};
\addlegendentry{\scriptsize{distance fusion}}
\addplot[very thick,color=black,dashed,mark=none] table [smooth,x=timestamp,y=danger_fusion_avg,col sep=comma]{all.csv};
\addlegendentry{\scriptsize{danger fusion}}
\addplot[thick,dotted,color=black,mark=none] table [smooth,x=timestamp,y=danger_tracker,col sep=comma]{all.csv};
\addlegendentry{\scriptsize{tracker}}
\draw[color=black!40,line width=0.25mm] (0,1) -- (5,1);
\end{axis}
\end{tikzpicture}
\caption{DF computed from distances recorded by the sensors (left) and as obtained by \emph{danger} and \emph{distance fusion} (right).}
\label{fig:dfs}
\end{figure}
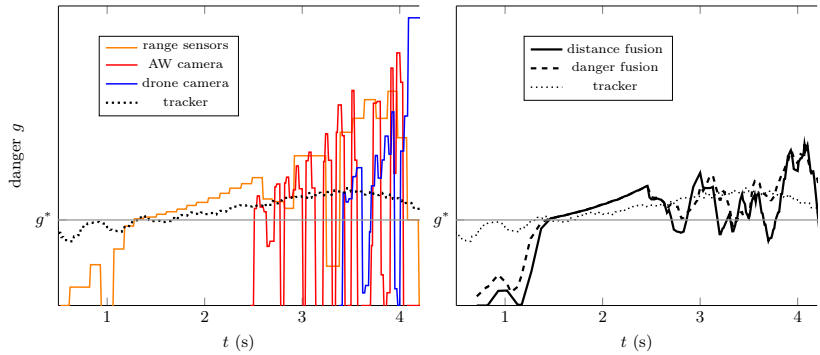

Similar considerations can be done for the \emph{danger fusion}. The DF values computed by the three sensors are displayed and compared with the tracker data in Fig.~\ref{fig:dfs} (left). Those values are very noisy and unstable, while the fusion offers a smoother output that is more suitable to danger evaluation, as depicted by Fig.~\ref{fig:dfs} (right).


For a quantitative evaluation, we computed a \emph{Root Mean Square Error} (RMSE) by comparing the DFs based on the tracker, with those obtained from the single sensors, and with the \emph{distance} and \emph{danger fusion} outputs. Such a descriptor cannot be computed for the \emph{voting fusion}, which returns a binary output. When coping with such outputs, we quantify the classification performance of the different methods in terms of accuracy, precision and recall. To interpret those measures, note that a \emph{false positive} is a case where a safe output is recognised as a dangerous situation. 
The overall results are displayed in Tab.~\ref{tab:rmse}. 


In our experimentation with single technologies, the range sensors provided the best performance in terms of RMSE, accuracy, and precision. However, the AW camera obtained the best recall. 
At the fusion level, the worst performance was given by the \emph{voting fusion}.
In spite of a higher RMSE compared to the \emph{danger fusion}, the accuracy of the 
\emph{distance fusion} was significantly higher. This confirms the idea that the best results are obtained by performing the fusion before any further signal processing \cite{hall}.


\begin{table}[htp]
\centering
\begin{tabular}{lp{1cm}rp{1cm}rrr}
\hline
Source/Fusion Technique&&RMSE&&Accuracy&Recall&Precision\\
\hline
Range sensors&&0.582&&{\bf 0.92}&0.89&{\bf 0.99}\\
AW camera&&0.948&&0.77&{\bf1.0}&0.59\\
Drone camera&&1.923&&0.51&0.63&0.32\\
\hline
Distance fusion&&0.447&&{\bf 0.92}&0.91&0.95\\
Danger fusion&&{\bf 0.361}&&0.83&0.78&0.96\\
Voting fusion&&-&&0.67&0.50&0.38\\
\hline
\end{tabular}
\caption{Performance evaluation of sensors and fusion techniques.}
\label{tab:rmse}
\end{table}

\section{Conclusions} \label{sec:conclusion}

In this paper, we have addressed the problem of safe road-crossing by autonomous wheelchairs supported by flying drones, using multi-sensor fusion. We have focused on the generation of a relevant laboratory dataset from multiple artificial vision and distance sensors installed on RoboMasters operating in ROS and OptiTrack enviroments. We made the dataset publicly available to the scientific community for further experimentation and performance evaluation. We also designed an analytical danger function to enable run-time risk assessment for road-crossing decision support. We have experimentally evaluated the danger function to provide some preliminary results as a proof-of-concept. The function is based on physical conditions and therefore represents a case of explainable decision fusion, whereas the output of some sensors, especially those based on artificial vision, is affected by a variable degree of opacity and uncertainty that need to be quantified to support probabilistic safety evaluation. Such an evaluation can be performed using an approach similar to the ones already introduced in  \cite{Flammini2020} and \cite{tas23}, and will be developed in future works. As future developments, we also plan to extend the dataset and experimentation with more complex real-world situations to be recognised and managed, such as multiple vehicles/obstacles from different directions, as well as with interference and disturbances due to vibrations, dirty camera lenses, glares, darkness, obstructions, weather conditions (rain, fog, etc.), some of which can be simulated by applying appropriate software filters. Although we focused on a specific application, we believe our contribution can be extended to other crossing scenarios, e.g., to support vision impaired people with similar assistive technologies.

\begin{credits}
\subsubsection{\ackname} This work was supported by the Swiss State Secretariat for Education, Research and lnnovation (SERI). 
The project has been selected within the European Union’s Horizon Europe research and innovation programme under grant agreement: HORIZON-CL4-2021-HUMAN-01-01. 
Views and opinions expressed are however those of the authors only and do not necessarily reflect those of the funding agencies, which cannot be held responsible for them.
\subsubsection{\discintname}
\end{credits}
\bibliographystyle{splncs04}
\bibliography{bibliography}
\end{document}